\documentclass[sigconf]{acmart}
\AtBeginDocument{%
  }

\usepackage{algorithmic}
\usepackage{algorithm}
\usepackage{graphicx}
\usepackage{textcomp}
\usepackage{xcolor}
\usepackage{booktabs}
\usepackage{subcaption}
\usepackage[table]{xcolor}
\usepackage{booktabs}

\setcopyright{none}
\begin{document}

\title{SensingAgents: A Multi-Agent Collaborative Framework for Robust IMU Activity Recognition}



\author{Naiyu Zheng}
\email{naiyzheng2-c@my.cityu.edu.hk}
\affiliation{%
  \institution{City University of Hong Kong}
  \country{Hong Kong}
}

\author{Tianlong Yu}
\email{tommyyu21@163.com}
\affiliation{%
  \institution{Hubei University}
  \country{China}
}

\author{Haochen Yin}
\email{1155178600@link.cuhk.edu.hk}
\affiliation{%
  \institution{The Chinese University of Hong Kong}
  \country{Hong Kong}
}

\author{Xiaoyi Fan}
\email{xiaoyi.fan@ieee.org}
\affiliation{%
  \institution{Shenzhen MSU-BIT University}
  \country{China}
}

\author{Xiping Hu}
\email{huxp@bit.edu.cn}
\affiliation{%
  \institution{Shenzhen MSU-BIT University}
  \country{China}
}

\author{Zhimeng Yin}
\email{zhimeyin@cityu.edu.hk}
\affiliation{%
  \institution{City University of Hong Kong}
  \country{Hong Kong}
}


\begin{abstract}
Human Activity Recognition (HAR) using Inertial Measurement Unit (IMU) sensors is a cornerstone of mobile health, smart environments, and human-computer interaction. However, current deep learning-based HAR models often struggle with heavy reliance on labeled data, position-specific ambiguity, and a lack of transparent reasoning. Inspired by the advanced agents framework, which emulates a collaborative agent using Large Language Models (LLMs), we propose SensingAgents, a novel multi-agent system for robust IMU activity recognition. SensingAgents organizes LLM-powered agents into specialized roles: an group of Analyst Agents for position-specific sensor analysis (arm, wrist, belt, pocket), a pair of Advocate Agents that resolves sensor conflicts through dynamic and static dialectical debates, and a Decision Agent that ensures reliability under sensor drift or failure. Evaluation on the Shoaib dataset demonstrates that SensingAgents significantly outperforms state-of-the-art single-agent and multi-agent LLM models, achieving an accuracy of 79.5\% in a zero setting—29\% higher than existing agent models and 9.4\% higher than deep learning baselines—particularly in complex scenarios where multi-sensor data is conflicting or noisy. Our work highlights the potential of multi-agent collaborative reasoning for advancing the robustness and interpretability of ubiquitous sensing systems.
\end{abstract}

\ccsdesc[500]{Human-centered computing~Ubiquitous and mobile computing systems and tools}

\keywords{Human Activity Recognition, Multi-Agent Systems, Large Language Models, Sensor Fusion, IMU Sensing}


\maketitle

\section{Introduction}
Human Activity Recognition (HAR) has emerged as a pivotal research area within the broader domain of ubiquitous computing and the Internet of Things (IoT). Its applications span a diverse range, from enhancing personalized healthcare and elder care systems to optimizing fitness tracking, facilitating smart home automation, and enabling immersive augmented reality experiences \cite{9055403, chiu2024active, zhang2022mobi2sense}. The pervasive deployment of wearable devices, including smartphones, smartwatches, and dedicated Inertial Measurement Unit (IMU) tags, has led to an unprecedented deluge of inertial data. This data encompasses acceleration, angular velocity, and magnetic field strength, providing a rich, continuous stream of information about human physiological and behavioral states. In essence, it serves as the digital heartbeat of modern health analytics, offering invaluable insights into the daily routines and well‑being of millions of individuals \cite{kim2019imu}.

\begin{figure}[!h]
\vspace{-0.1in}
  \centering
  \includegraphics[width=0.7\linewidth]{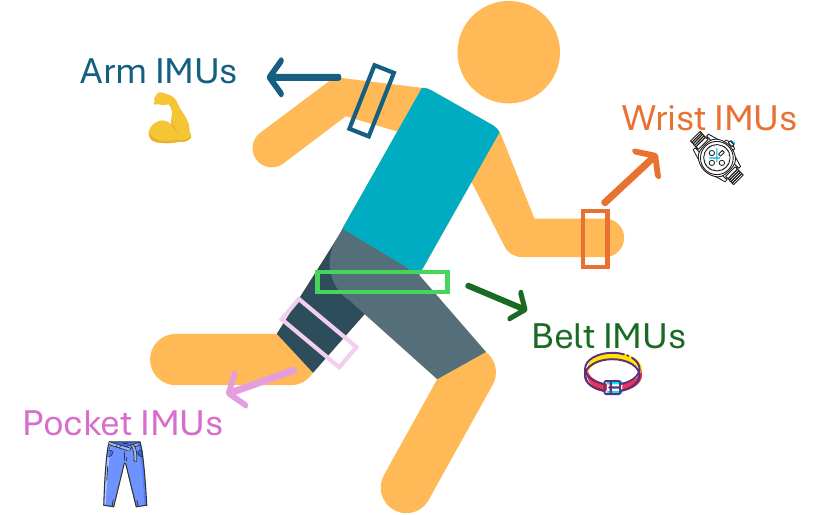}
  \vspace{-5pt}
  \caption{The IMU data is normally collected at several body positions to capture a comprehensive representation of human activity, each with distinct motion dynamics.}
  \label{fig:MultipleIMUs}
  \vspace{-2.5pt}
\end{figure}

However, despite the abundance of IMU data, robust HAR remains challenging along three fundamental dimensions: \textbf{interpretability}, \textbf{multi-position adaptability}, and \textbf{dependence on labeled data}. First, most high-performing HAR systems rely on deep learning architectures such as Convolutional Neural Networks (CNNs), Recurrent Neural Networks (RNNs), and more recently Transformers, which are effective at modeling spatial and temporal patterns in sensor streams \cite{ordonez2016deep, 10.1145/3711896.3737226}. Yet these models typically behave as ``black boxes,'' offering little insight into why a specific prediction was made. This lack of \textbf{interpretability} is particularly problematic in high-stakes applications such as remote patient monitoring and fall detection, where users and practitioners must understand the rationale behind a classification in order to trust, debug, and refine the system.

Second, HAR systems must operate under substantial variability across body positions. In real deployments, sensors placed on the arm, wrist, belt, or pocket capture the same activity through different motion signatures, and those signatures may even conflict. For example, during walking, a pocket-mounted sensor may strongly reflect locomotion, while a wrist-mounted sensor may appear nearly static if arm swing is limited. Traditional sensor fusion methods, including early fusion and late fusion \cite{demirel2025using, tang2023comparative}, often merge these signals without explicitly reasoning about why they disagree. As a result, they struggle with \textbf{multi-position adaptability}, especially when placements are heterogeneous, noisy, or partially unreliable.

Third, most supervised HAR pipelines require substantial labeled datasets to achieve strong performance. Collecting such annotations across users, activities, device placements, and environmental conditions is costly and difficult to scale. This heavy \textbf{labeled data requirement} limits deployment in realistic settings, where new sensor configurations or user populations may not have sufficient annotated examples. Although recent LLM-based sensing methods have opened the door to zero-shot and few-shot activity recognition \cite{li2025sensorllm, ji2024hargpt, xu2024penetrative, yan2025mmexpert}, existing approaches usually adopt a single-agent design that processes all sensor streams jointly. This can lead to hallucinated reasoning, weak grounding in raw signal characteristics, and poor handling of conflicts across positions. More recent multi-agent frameworks \cite{le2025multi, zhao2025multiagentscr, yoon2026consensus}, improve collaborative reasoning but still leave two important limitations insufficiently addressed: \textbf{high token consumption} and \textbf{weak conflict resolution}. Because multiple agents repeatedly exchange long natural-language summaries over the full sensor context, inference cost grows quickly, making such frameworks difficult to deploy at scale. At the same time, when heterogeneous body-position sensors provide contradictory evidence, these methods typically rely on semantic discussion or shallow consensus mechanisms rather than an explicit debate structure grounded in signal-level analysis. As a result, they may still struggle to determine whether disagreement reflects true activity dynamics, sensor noise, or position-specific ambiguity.

To address these three challenges, we propose \textbf{SensingAgents}, a modular multi-agent framework for robust IMU activity recognition. For \textbf{interpretability}, SensingAgents decomposes inference into explicit analyst, debate, and decision stages, enabling the system to produce transparent intermediate rationales rather than a single opaque prediction. For \textbf{multi-position adaptability}, the framework assigns position-specific Analyst Agents to independently interpret local sensor streams from the arm, wrist, belt, and pocket, then uses Dynamic and Static Advocate Agents to debate and reconcile conflicting evidence across placements. For \textbf{reducing reliance on labeled data}, SensingAgents leverages the zero-shot reasoning ability of LLMs while grounding each agent with tailored tools such as Fast Fourier Transform (FFT), peak analysis, and spectral density estimation, so that predictions remain tied to measurable signal properties instead of large amounts of task-specific supervision. This design also addresses the limitations highlighted above: by assigning each agent a focused role, SensingAgents avoids unnecessary full-context exchanges that inflate token consumption, and by introducing an explicit Dynamic--Static debate followed by a Decision Agent, it resolves cross-position conflicts more systematically than shallow semantic consensus. Together, these components form a ``society of agents'' that combines structured reasoning, token-efficient collaboration, and signal-level grounding to deliver more trustworthy and robust HAR.

The main contributions of this work are summarized as follows:
\begin{itemize}
    \item We propose SensingAgents, the first position-specific multi-agent LLM framework for HAR that enhances both the accuracy and the transparency of activity recognition.
    \item We introduce a modular tool-calling layer that bridges LLM reasoning with deterministic signal analysis on raw data, ensuring physical interpretability and token efficiency.
    \item We design a movement-specific debate mechanism that resolves conflicts between heterogeneous body positions using a ''Dynamic-Static'' reasoning logic.
    \item We validate the framework on the Shoaib dataset, demonstrating state-of-the-art performance in zero-shot adaptation and robustness against signal corruption.
\end{itemize}

\vspace{-0.10in}
\section{Motivation}

The motivation for SensingAgents is rooted in the empirical performance gaps observed when applying conventional HAR models to multi-position sensing scenarios. 

\vspace{-0.10in}
\subsection{Positional Sensitivity and Domain Shift}

The variability of IMU signals across different body parts is a fundamental challenge in HAR. Sensors positioned at the user's center of gravity (e.g., the belt) provide stable, torso-centric data that is highly informative for identifying user-independent locomotion patterns. Conversely, sensors on the extremities, such as the wrist or arm, are susceptible to additional noise from non-ambulatory gestures like drinking, typing, or expressive hand movements. 

\begin{table}[htbp]
\caption{Positional Signal Characteristics}
\label{tab:positional_reliability}
\centering
\small  
\setlength{\tabcolsep}{4pt}  
\begin{tabular}{lp{1.7cm}cl}
\toprule
\textbf{Position} & \textbf{Dominant Motion} & \textbf{Reliability (Walk)} & \textbf{Reliability (Sit)} \\
\midrule
Arm & Rotational/Swing & High & Low (Arm-shift noise) \\
Belt & Vertical/Oscillatory & Very High & Very High (Gravity) \\
Pocket & Linear/Impact & High & Medium (Clothing) \\
Wrist & Multi-axial & Medium & Low (Gesticulation) \\
\bottomrule
\end{tabular}
\end{table}

Table \ref{tab:positional_reliability} compares the typical signal characteristics and classification reliability for common activities across these positions, illustrating why a one-size-fits-all model struggles to generalize. The data implies that while the belt is optimal for core activity detection, the wrist and arm provide complementary information for detecting upper-body engagement. Monolithic models often fail to capture these nuances because they either aggregate all features into a flat vector or rely on fixed input shapes that cannot adapt if a sensor is removed or added. This motivates a modular architecture where each position is treated as a specialized analytical domain.

\begin{table*}[t]
\caption{Comparison of representative models for HAR}
\label{tab:model_comparison}
\centering
\footnotesize  
\setlength{\tabcolsep}{8pt}
\begin{tabular}{lllccc}
\toprule
\textbf{Method} & \textbf{Category} & \textbf{Accuracy} & \textbf{Interpretability} & \textbf{Multi-position Adaptability} & \textbf{Labelled Data Requirements} \\
\midrule
SVM, Random Forest & ML-based & High & Low & Low & Medium \\
DeepConvLSTM & DL-based & High & Low & Low & High \\
LIMU-BERT & Self-supervised & High & Low & Low & High \\
SensorLLM & Single-Agent & Medium & Medium & Low & Low \\
MultiAgentsCR & Multi-Agent & Medium & Medium & High & Low \\
\textbf{SensingAgents (Ours)} & Multi-Agent & High & High & High & Zero \\
\bottomrule
\end{tabular}
\end{table*}

\vspace{-0.10in}
\subsection{Preprocessing and Cognitive Load Mitigation for LLM-based Sensing}
Directly applying LLMs to raw sensor data necessitates careful consideration of token limits and input dimensionality. High-frequency sensors, such as an IMU sampled at 100 Hz, produce thousands of data points within a short window, which can easily overwhelm the context window of LLMs. Furthermore, when relying on a single LLM agent to process all concurrent sensor streams, the model frequently experiences ''cognitive overload.'' Confronted with a deluge of numerical observations and textual descriptions from multiple body positions, a single-agent LLM may struggle to synthesize contradictory evidence and is therefore prone to hallucinations \cite{alansari2025large}, producing plausible but incorrect activity classifications by over-indexing on noisy signals or neglecting conflicting information.

\begin{figure}[ht]
    \centering
    \includegraphics[width=0.65\linewidth]{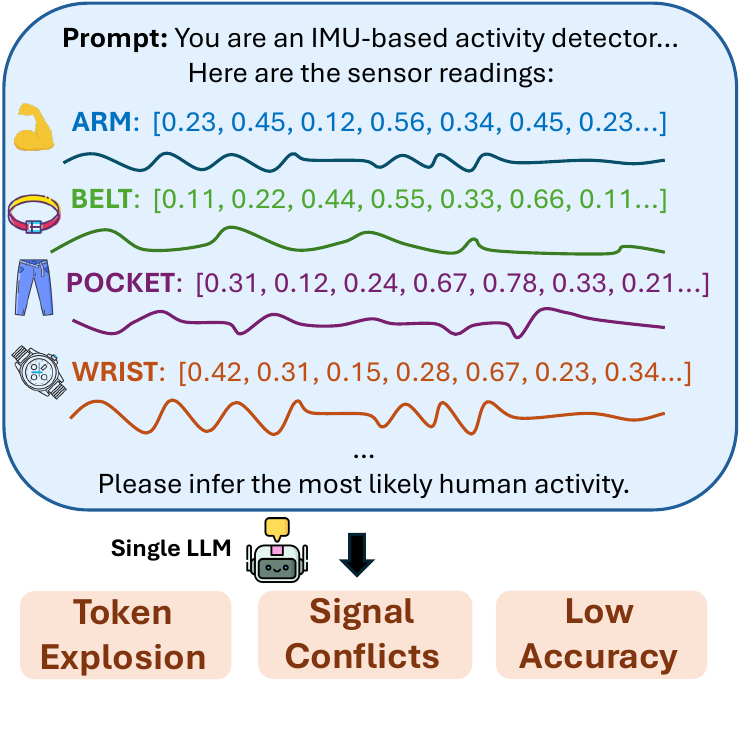}
    \vspace{-13pt}
    \caption{A traditional single agent leads to cognitive overload and hallucinations.}
    \label{fig:SingleAgent}
    \vspace{-2.5pt}
\end{figure}

To mitigate input complexity, existing approaches typically employ downsampling and quantization strategies. For instance, a 360 Hz signal can be downsampled to 36 Hz and quantized to integer values, reducing sequence length while preserving essential morphological features for activity recognition. However, even with such token optimization, relying solely on an LLM’s internal textual reasoning to interpret complex signal morphology remains suboptimal. These limitations motivate the development of structured agentic architectures that distribute reasoning across multiple agents, resolve conflicts among heterogeneous sensor observations, and invoke external signal processing tools to verify and refine sensing logic.

\vspace{-0.10in}
\subsection{Transparent Interpretability}

A major barrier to the adoption of AI in monitoring and analysis is the 'black-box' nature of deep neural networks. In a geriatric care scenario, a system must not only detect a fall but also provide the reasoning behind the alert to minimize false alarms and build trust with medical staff. Existing literature emphasizes the need for Explainable AI \cite{jeyakumar2023x, abdelaal2024exploring} that can articulate human-intuitive rationale. 

However, simple post-hoc explanations generated by deep learning models often lack causal transparency. To address this limitation, LLMs can serve as as the reasoning core within a multi-agent framework to achieve \textit{Transparent Interpretability}. The system is expected to generate a clear Chain-of-Thought (CoT) \cite{wei2022chain} that explicitly documents how positional data are analyzed, how conflicting evidence is debated, and how the final consensus is reached. This transparent reasoning process allows users to trace the system's decision logic, verifying whether a ``walking'' classification arises from a legitimate gait pattern or from anomalous arm movements. Such transparency is essential for ensuring safety, accountability, and user trust in ubiquitous sensing applications.

\section{Related Work}
The field of HAR has evolved through several distinct phases, from traditional machine learning and deep learning to the recent emergence of LLMs and multi-agent systems, as shown in Table \ref{tab:model_comparison}.

\vspace{-0.10in}
\subsection{Traditional and Deep Learning for HAR}

Early Human Activity Recognition (HAR) systems used a two-stage pipeline: extracting hand-crafted features from IMU time-series (e.g., mean, variance, FFT-based frequency features) \cite{zhou2007inertial}, then classifying them with SVMs, Random Forests, or HMMs \cite{mannini2010machine}. While effective in controlled settings, these methods relied on domain expertise and often failed to generalize across users or environments. To overcome these generalization challenges, deep learning has been increasingly adopted in HAR, enabling automatic feature extraction from raw sensor data. For instance, CNNs capture spatial patterns in multi-channel IMU signals, and RNNs model temporal dependencies \cite{ordonez2016deep}. These models outperformed traditional approaches on benchmarks like the Shoaib dataset \cite{shoaib2014fusion}. More recently, Transformer-based architectures leverage self-attention to capture long-range temporal and cross-sensor dependencies \cite{10.1145/3711896.3737226}, while self-supervised frameworks such as LIMU-BERT \cite{xu2021limu} shows promise by pre-training on large unlabeled IMU data.

Despite progress, deep models remain largely "black box" systems, which limits interpretability which is critical in safety-sensitive domains like healthcare. They also struggle with sensor noise, drift, and inter-user variability, requiring heavily labeled data and transfer learning. Monolithic architectures still lack explicit mechanisms to handle conflicting signals from multiple body-worn sensors.

\begin{figure*}[htbp]
    \centering
    \includegraphics[width=0.7\textwidth]{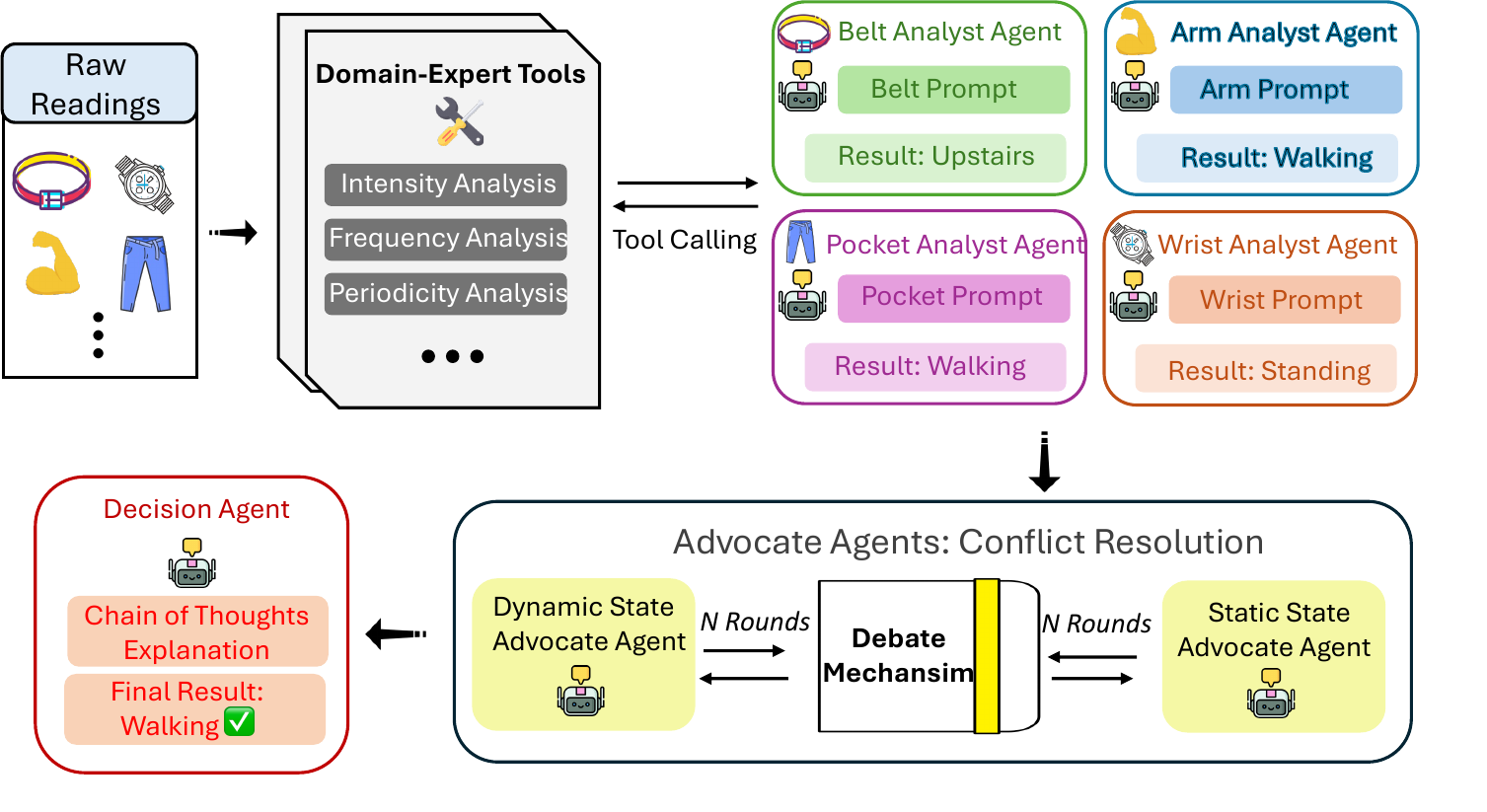}
    \vspace{-13pt}
    \caption{Workflow of SensingAgents.}
    \label{fig:workflow}
\end{figure*}

\vspace{-0.10in}
\subsection{LLMs in Ubiquitous Sensing}
Inspired by LLM success in NLP, researchers have applied them to non-linguistic data like IMU streams \cite{ferrara2024large, an2026iot, asif2025llasa}. LLMs' ability to process and reason with text makes them well suited for interpreting sensor data and generating human-readable explanations. Projects such as SensorLLM \cite{li2025sensorllm}, HAR-GPT \cite{ji2024hargpt}, and ZARA \cite{li2025zara} pioneered this direction. These works show that LLMs can perform zero-shot or few-shot activity recognition by aligning pre-processed sensor data (e.g., symbolic representations or summaries) with natural language descriptions of activities. For example, an LLM prompted with a wrist accelerometer summary like "high frequency, moderate amplitude oscillations" can infer the activity. Recent efforts also explore LLMs for late multimodal fusion \cite{demirel2025using} and semantic alignment \cite{yan2025large}. However, most current approaches use a single-agent architecture, where one LLM processes all data. While effective for clean signals, this setup has key limitations: cognitive overload with multi-modal, multi-position data, difficulty reconciling conflicting or noisy inputs, and vulnerability to sensor drift and spurious signals due to lack of specialization.

Given these limitations of single-agent models, researchers have recently turned to multi-agent architectures as a potential solution. For example, GLOSS \cite{le2025multi} utilizes a multi-agent LLM network for suggesting and correcting human activity annotations by triangulating self-reports and passive sensing data. MultiAgentsCR \cite{zhao2025multiagentscr} introduces a collaborative reasoning framework that utilizes multiple role-playing agents to collaboratively analyze textual descriptions of human activities and simulate conference-style discussions for activity recognition. ConSensus \cite{yoon2026consensus} proposes a training-free multi-agent collaboration framework that decomposes multimodal sensing tasks into specialized, modality-aware agents. Despite these advancements, existing multi-agent sensing frameworks still exhibit notable limitations. They primarily focus on textual reasoning without invoking deterministic signal processing tools, lack explicit conflict resolution mechanisms for heterogeneous sensor placements, and assume fixed known positions, making them brittle to variations across users or deployment contexts.



\vspace{-0.10in}
\section{Design: The SensingAgents Architecture}

The SensingAgents framework is designed as a hierarchical, multi-layered system, each layer populated by specialized LLM-powered agents shown in Figure~\ref{fig:workflow}. This modular architecture decomposes complex IMU sensing into discrete stages, including raw signal acquisition and analysis, sophisticated conflict resolution, and the ultimate assessment. The system ensures that each task is handled by an optimized agent to achieve high accuracy and robust interpretability.

\vspace{-0.10in}
\subsection{The Analyst Agent: Position-Specific Expertise and Tool Integration}

The foundational layer of SensingAgents consists of position-specific analyst agents, including Arm, Wrist, Belt, and Pocket Analysts, each assigned to a distinct sensor position. These agents are responsible for preprocessing and interpreting raw IMU data from their respective locations, leveraging specialized tools and domain knowledge to extract motion features.

\subsubsection{Agent Personas and Specializations}
Each analyst agent is endowed with a distinct persona and a set of specialized capabilities tailored to the kinematic characteristics typically observed at its assigned body position:

\textbf{Arm Analyst}: This agent monitors the upper arm to recognize coarse-grained movements such as lifting, reaching, and pushing. It distinguishes static postures (e.g., sitting) from active engagement (e.g., biking) by analyzing spatial orientation and motion trajectories. Technically, it decomposes raw signals into rotational and translational components, employing gravity compensation to isolate dynamic linear acceleration from gravitational artifacts, thereby capturing the underlying force of large-scale limb actions.

\textbf{Wrist Analyst}: Positioned for fine-grained gesture recognition, this agent captures high-frequency hand movements and the rhythmic arm swings characteristic of locomotion. It specializes in identifying periodic patterns—such as walking, jogging, or typing—by transforming time-series data into the frequency domain. Utilizing techniques like Fast Fourier Transform (FFT), it extracts dominant frequency peaks (e.g., $1.5$–$2.5$ Hz for gait) and spectral energy to quantify the cadence and intensity of repetitive activities.

\textbf{Belt Analyst}: Positioned near the body's center of mass (e.g., on the hip or waist), this agent serves as the primary reference for gait analysis and postural stability. It excels at distinguishing static postures from dynamic transitions by monitoring torso displacement and orientation. Utilizing statistical features such as mean acceleration magnitude and zero-crossing rates, the Belt Analyst characterizes movement periodicity and provides critical data for detecting high-impact events like falls or sudden changes in body alignment.

\textbf{Pocket Analyst}: This agent focuses on lower-limb kinematics from a thigh-placed device, specializing in leg-driven activities such as cycling and stair negotiation. By correlating acceleration patterns with gravitational vectors, it infers changes in elevation and leg propulsion mechanics. It can differentiate between the distinct kinematic signatures of ascending versus descending stairs, for instance, by analyzing the phase relationships between vertical acceleration and angular velocity.

\subsubsection{Domain-Expert Tool Design: Bridging Numerical Precision and Semantic Reasoning}

A fundamental limitation of applying LLMs to sensor data is the \textit{modality gap}: LLMs excel at symbolic reasoning over natural language but lack the capacity to perform reliable numerical computation on raw time-series. Our ablation study confirms this limitation empirically—removing tool-calling drops accuracy from 79.5\% to 60.7\%, a degradation of nearly 20 percentage points. This motivates our central design principle: \textbf{decouple numerical feature extraction from semantic reasoning} by equipping each analyst agent with a deterministic \textit{Domain-Expert Tools} accessible via tool-calling.

Unlike the code-generation method, where the LLM autonomously writes and executes signal processing code (frequently producing flawed pipelines or infinite loops, see Section~Evaluation), our tools are \textit{human-designed, pre-verified} modules grounded in domain expertise from HAR research. Each tool transforms a raw IMU window $\mathbf{x} \in \mathbb{R}^{T \times C}$ ($T$ samples, $C$ channels) into a compact \textit{semantic feature descriptor} that the LLM can reason about in natural language. The toolbox is organized into three functional tiers based on discriminative purpose.

\noindent\textbf{Tier~1: Activity-Level Discrimination (Static vs.\ Dynamic).}
The first tier addresses the coarsest classification boundary. The \textit{Signal Magnitude Area} (SMA) provides a scalar measure of overall motion intensity:
\begin{equation}
\text{SMA}_\text{gyro} = \frac{1}{T}\sum_{t=1}^{T}\left(|\omega_{t,x}| + |\omega_{t,y}| + |\omega_{t,z}|\right)
\end{equation}

Based on the observed data distribution, $\text{SMA}_\text{gyro}$ exhibits a clear separation between activity classes: static activities yield very low values, whereas dynamic activities produce substantially larger ones. This distinct contrast enables the LLM agent to immediately narrow the hypothesis space. The complementary \textit{Statistical Feature Extraction} tool computes per-channel mean $\mu_c$, standard deviation $\sigma_c$, and RMS alongside vector magnitudes $\|\mathbf{a}_t\|$ and $\|\boldsymbol{\omega}_t\|$, providing the agent with a quantitative ``vital sign'' of each sensor position. This tier eliminates the need for the LLM to mentally estimate variance from raw numerical sequences—a task at which language models demonstrably fail.

\begin{figure}[!h]
    \centering
    \begin{subfigure}{\columnwidth}
        \centering
        \includegraphics[width=0.8\linewidth]{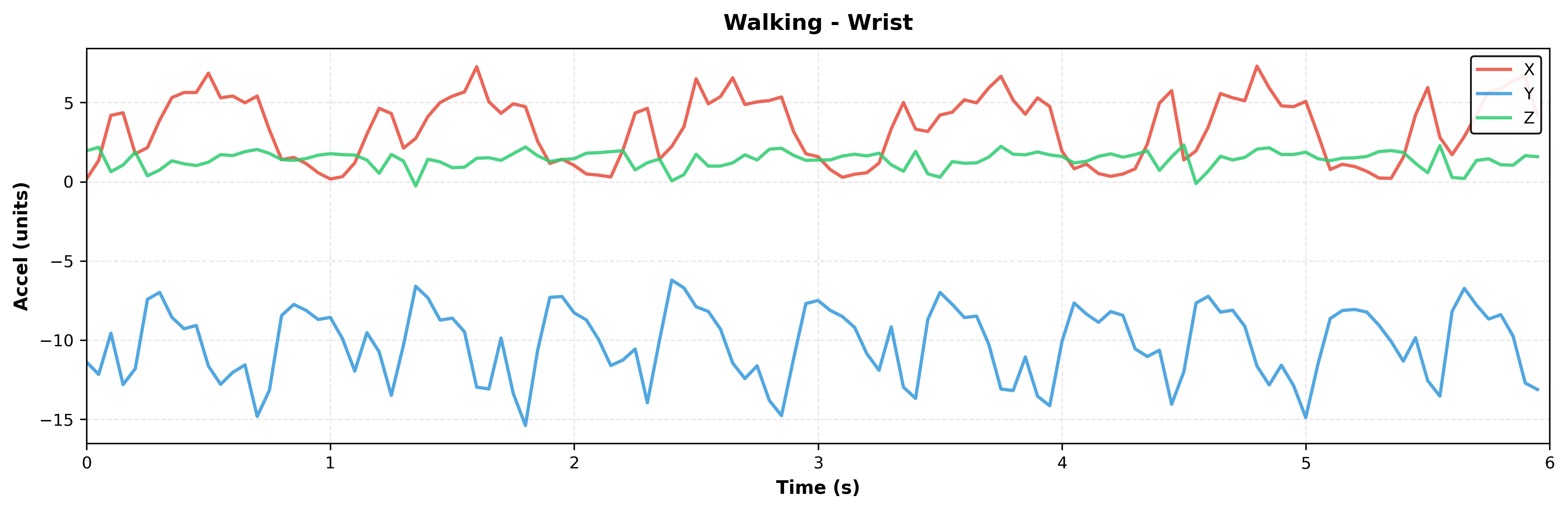}
        \caption{The accelerometer readings on wrist when walking.}
        \label{fig:user0_wrist_accel_walking}
    \end{subfigure}
    
    
    \begin{subfigure}{\columnwidth}
        \centering
        \includegraphics[width=0.8\linewidth]{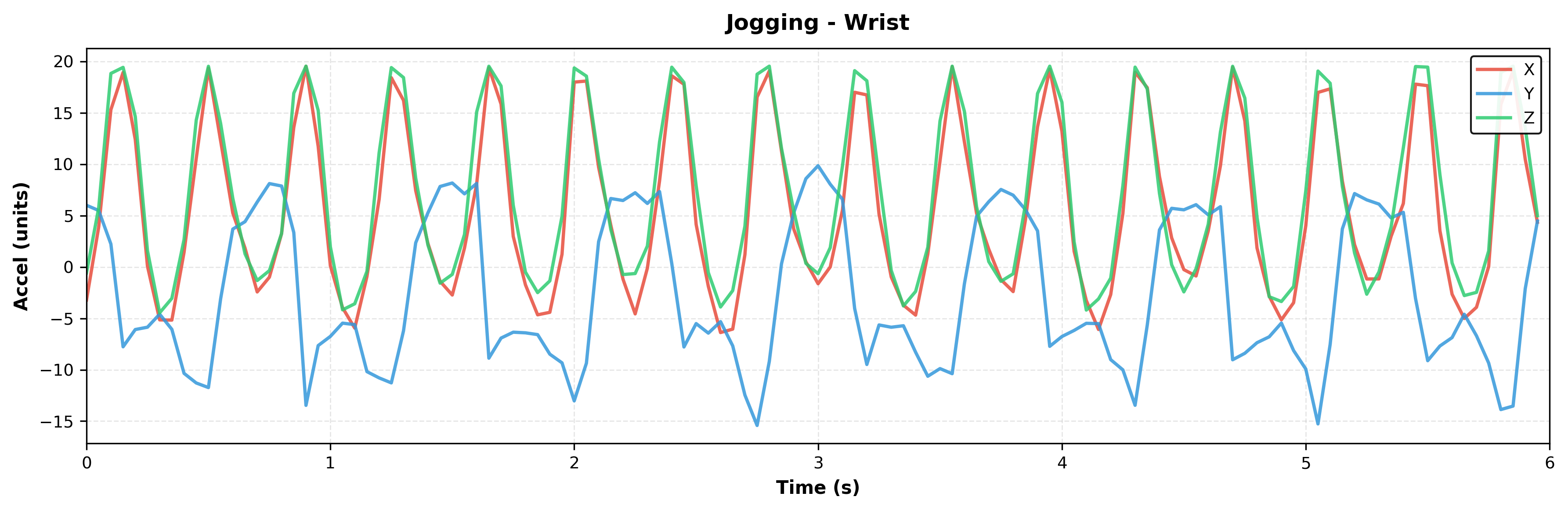}
        \caption{The accelerometer readings on wrist when jogging.}
        \label{fig:user0_wrist_accel_jogging}
    \end{subfigure}
    \caption{Even though jogging and walking are rhythmic dynamic movements, agents can make comprehensive judgments based on the amplitude of accelerometers at specific locations without inputting the entire time series. As shown in the figure, the amplitude and frequency of jogging are higher than walking, and our tailored tool can capture this information.}%
    \label{fig:user0_wrist_accel}
\end{figure}

\noindent\textbf{Tier~2: Cadence and Periodicity Analysis.}
Once dynamic activity is established, the second tier disambiguates locomotion categories. The \textit{FFT Analysis} tool applies the Discrete Fourier Transform:
\begin{equation}
X_c[k] = \sum_{t=0}^{T-1} x_{t,c}\, e^{-j2\pi kt/T}, \quad k = 0, 1, \ldots, \lfloor T/2 \rfloor
\end{equation}
suppresses the DC component, and returns the top-3 dominant frequencies $\{f_k^*\}$ with amplitudes and spectral energy $E_c = \sum_k |X_c[k]|^2$. The dominant frequency directly encodes gait cadence, with jogging normally exhibiting a higher dominant frequency than walking. The spectral energy ratio between the low band (${\leq}2.5$\,Hz) and high band ($>2.5$\,Hz) further separates rhythmic locomotion from irregular motion.

Complementarily, the \textit{Step Detection} tool identifies prominent peaks in the vertical acceleration $a_y$ and computes the step frequency and regularity:
\begin{equation}
\hat{f}_\text{step} = \frac{1}{\bar{\Delta\tau}}, \quad
\text{regularity} = 1 - \frac{\sigma_{\Delta\tau}}{\bar{\Delta\tau}}
\end{equation}
where $\bar{\Delta\tau}$ is the mean inter-peak interval and $\sigma_{\Delta\tau}$ its standard deviation. The \textit{Autocorrelation Periodicity} tool validates these findings through the normalized autocorrelation:
\begin{equation}
R(\ell) = \frac{\sum_{t=0}^{T-1-\ell}(a_{t,y}-\bar{a}_y)(a_{t+\ell,y}-\bar{a}_y)}{\sum_{t=0}^{T-1}(a_{t,y}-\bar{a}_y)^2}
\end{equation}
The first peak $R(\ell^*) > 0.2$ yields the stride period, which the LLM can cross-reference with FFT results. This dual-verification design is critical: the ablation without tool-calling shows that LLMs, when forced to infer periodicity from raw numbers alone, frequently hallucinate cadence values.

\noindent\textbf{Tier~3: Fine-Grained Disambiguation.}
The most challenging classification boundaries (e.g., upstairs vs.\ downstairs) require features that encode subtle biomechanical asymmetries. The \textit{Vertical Asymmetry Analysis} tool computes three discriminative quantities from the vertical acceleration $a_y$:
\begin{equation}
\gamma_y = \frac{1}{T}\sum_{t=1}^{T}\left(\frac{a_{t,y} - \mu_y}{\sigma_y}\right)^3, \quad 
\eta = \frac{E^+}{E^+ + E^-}
\end{equation}
where $\gamma_y$ is the skewness (upstairs: $\gamma_y > 0$, positive-dominant push; downstairs: $\gamma_y < 0$, impact-dominant) and $\eta$ is the positive-to-total energy ratio. A peak-valley amplitude ratio $\rho_\text{pv} = \bar{A}^+ / \bar{A}^-$ captures the asymmetry in vertical impacts. These features are inaccessible to pure LLM reasoning because they require precise numerical integration over the acceleration distribution.


\begin{figure}[!h]
  \centering
  \includegraphics[width=0.95\linewidth]{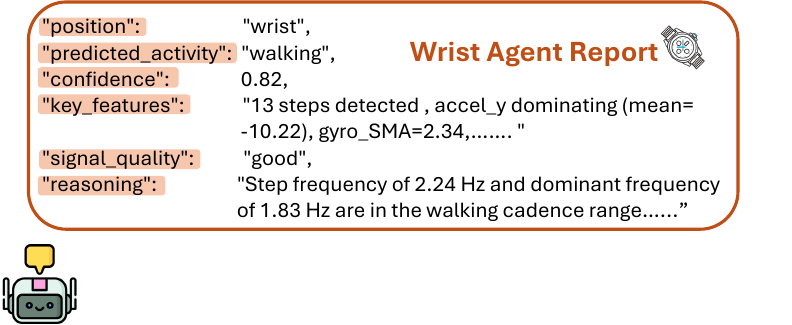}
  \caption{The wrist analyst agent report, which will be sent to the downstream agents for further analysis.}
  \label{fig:wrist_report}
\end{figure}

Upon completing the analysis, each analyst agent emits a structured report, as shown in Fig~\ref{fig:wrist_report}. This structured interface encodes both quantitative evidence and natural-language Chain-of-Thought rationale, decoupling position-specific sensing from downstream conflict resolution. The Advocate Agents receive not raw numbers but semantically grounded feature summaries, a representation that directly supports dialectical debate.


\vspace{-0.10in}
\subsection{The Advocate Agent: Dialectical Debate for Conflict Resolution}
The Advocate Agent constitutes the second layer of SensingAgents, tasked with resolving the inter-position conflicts that arise when analyst reports disagree. Rather than relying on simple majority voting, the system employs a structured adversarial debate mechanism between two opposing advocate agents.

\subsubsection{The Dynamic and Static Debate Protocol}
When the Analyst Team's reports exhibit inconsistencies or present conflicting activity classifications, for example, if the Wrist Analyst suggests ``Walking'' (due to arm swing) while the Belt Analyst suggests ``Standing'' (due to minimal torso movement), the Advocate Agents initiate a dialectical debate. This process involves two specialized researcher agents:

\begin{itemize}
    \item \textbf{Dynamic Activity Advocate}: This agent is programmed to advocate for a more dynamic, higher-intensity activity state. Its persona is to seek evidence that supports movement, change, and higher energy expenditure. It reviews all Analyst Reports, prioritizing those that indicate significant motion, higher magnitudes of acceleration, or faster angular velocities. It will construct arguments emphasizing the most active sensor readings and attempt to contextualize or downplay contradictory static readings (e.g., arguing that a static belt reading might be due to a brief pause during a dynamic activity).

    \item \textbf{Static Activity Advocate}: Conversely, this agent is tasked with advocating for a more static or lower-intensity activity state. Its persona is to identify and emphasize evidence that suggests minimal movement, stability, or lower energy expenditure. It scrutinizes Analyst Reports for indications of stillness, low acceleration magnitudes, or consistent orientation. It will construct arguments highlighting static sensor readings and attempt to explain away dynamic readings as artifacts, noise, or non-activity-related movements (e.g., attributing a wrist sensor's dynamic reading to fidgeting rather than locomotion).
\end{itemize}

\begin{figure}[!h]
  \centering
  \includegraphics[width=\linewidth]{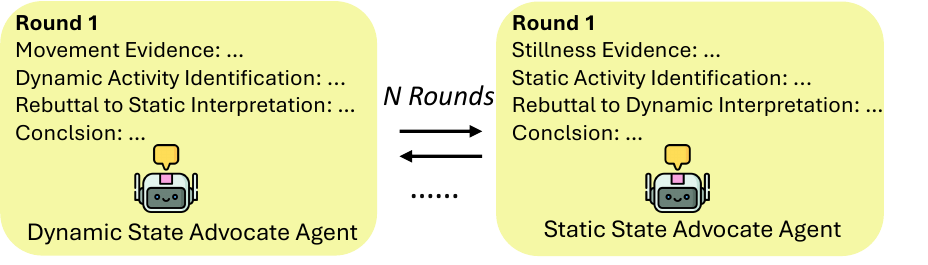}
  \caption{A round debate example of dynamic and static advocate agent.}
  \label{fig:debate}
\end{figure}

These two agents engage in a structured, multi-round discourse to reconcile divergent sensor interpretations:
\begin{enumerate}
    \item \textbf{Initial Stance}: Each advocate agent presents its initial activity hypothesis and the strongest supporting evidence from the Analyst Reports.
    \item \textbf{Evidence Presentation}: Agents take turns presenting additional evidence, counter-arguments, and justifications, drawing upon the detailed features and confidence scores provided in the Analyst Reports. They can also request further analysis from the Analyst Team if specific data points require deeper scrutiny.
    \item \textbf{Cross-Examination}: Agents actively challenge each other's interpretations, probing for weaknesses in the presented evidence or alternative explanations for observed sensor patterns. For instance, the Dynamic Advocate might ask the Static Advocate to explain a sudden spike in vertical acceleration if advocating for 'Standing'.
    \item \textbf{Rebuttal and Refinement}: Based on the cross-examination, agents refine their arguments, acknowledge limitations, or introduce new supporting evidence. This iterative process continues for a predetermined number of rounds or until a clear convergence or divergence of opinion is established.
\end{enumerate}

This dialectical process ensures rigorous multi-perspective appraisal, mitigating biases from localized or misleading sensor signals. By replacing black-box statistical aggregation with transparent, cross-validated reasoning, the framework significantly bolsters both the robustness and interpretability of the HAR system.

\vspace{-0.10in}
\subsection{The Decision Agent: The Ultimate Arbiter}
The Decision Agent serves as the definitive arbiter within the SensingAgents framework, functioning as a critical safeguard for system reliability. Its role is indispensable in high-stakes domains—such as mobile health monitoring, geriatric care, and human-robot collaboration, where erroneous activity classifications can have severe consequences. By performing a multi-criteria synthesis of the Analyst reports and the Advocate's dialectical outcomes, this agent ensures that final determinations are grounded in physically coherent evidence rather than isolated signal patterns.

Specifically, the Decision Agent is tasked with resolving ambiguity when the adversarial debate fails to yield a stable consensus or when confidence metrics remain below a predefined threshold. It proactively intervenes to weigh conflicting evidence, such as reconciling high-intensity limb movements with stationary torso data, to prevent premature or erroneous inferences. Beyond producing a discrete activity label, the agent generates a comprehensive diagnostic output comprising a quantitative confidence score and a natural language "Chain of Thought" (CoT) explanation. This explanation systematically decomposes the underlying reasoning by citing specific feature alignments and cross-agent validation results. Such transparency transforms the traditionally "black-box" sensing process into an interpretable narrative, fostering the auditability and user trust essential for seamless deployment in real-world assistive technologies.

\vspace{-0.10in}

\subsection{End-to-End Inference Pipeline}


The SensingAgents pipeline formalizes the transition from raw IMU signals to interpreted activity labels through a three-stage hierarchical process. Let $\mathcal{P} = \{p_i\}_{i=1}^N$ denote the $N$ sensor positions and $\mathbf{x}^{(i)} \in \mathbb{R}^{T \times C}$ the corresponding raw data windows. The activity label space is $\mathcal{Y}$. Let $\mathcal{L}$ denote the underlying LLM and $\mathcal{T}$ the signal processing toolbox. The inference procedure, detailed in Algorithm~\ref{alg:sensingagents}, is structured as follows:

\begin{algorithm}[tbp]
\caption{SensingAgents Inference Pipeline}
\label{alg:sensingagents}
\begin{algorithmic}[1]
\renewcommand{\algorithmicrequire}{\textbf{Input:}}
\renewcommand{\algorithmicensure}{\textbf{Output:}}
\REQUIRE Raw IMU windows $\{\mathbf{x}^{(i)}\}_{i=1}^{N}$, toolbox $\mathcal{T}$, LLM $\mathcal{L}$
\ENSURE Predicted activity $\hat{y}$, confidence $\hat{c}$, reasoning chain $\mathcal{E}$
\STATE \textit{/* Stage 1: Position-Specific Analysis (parallel) */}
\FOR{$i = 1$ \TO $N$ \textbf{in parallel}}
    \STATE Initialize analyst agent $\mathcal{A}_i$ with position-specific prompt
    \STATE $r_i (\hat{y}_i, c_i, \mathbf{f}_i, e_i) \leftarrow \mathcal{L}_\text{ReAct}(\mathcal{A}_i, \mathbf{x}^{(i)}, \mathcal{T})$ \COMMENT{Iterative feature extraction \& reporting}
\ENDFOR
\STATE \textit{/* Stage 2: Dialectical Debate */}
\STATE $h_0 \leftarrow \{r_1, \ldots, r_N\}$ \COMMENT{initial evidence}
\FOR{$j = 1$ \TO $R$}
     \STATE $d_j^{+} \leftarrow \mathcal{L}(\text{DynAdv}, h_{j-1})$ ; $d_j^{-} \leftarrow \mathcal{L}(\text{StaAdv}, h_{j-1}, d_j^{+})$ 
    \STATE $h_j \leftarrow h_{j-1} \cup \{d_j^{+}, d_j^{-}\}$
\ENDFOR
\STATE \textit{/* Stage 3: Decision Synthesis */}
\STATE $\hat{y}, \hat{c}, \mathcal{E} \leftarrow \mathcal{L}(\textsc{Advocate}, \{r_i\}, h_R)$
\RETURN $\hat{y}, \hat{c}, \mathcal{E}$
\end{algorithmic}
\end{algorithm}






\textbf{Stage 1: Position-Specific Analysis.} Each sensor position is assigned an independent analyst agent $\mathcal{A}_i$ operating in parallel. We employ a ReAct prompting strategy, allowing agents to iteratively invoke signal processing tools $\mathcal{T}$ to extract discriminative features. This stage culminates in $N$ structured reports $r_i$, which encapsulate local predictions $\hat{y}_i$, confidence $c_i$, key feature vector $\mathbf{f}_i$, and the underlying physical evidence $\mathbf{e}_i$.

\textbf{Stage 2: Dialectical Debate.} implements the Dynamic-Static dialectical debate. At each round $j$, the Dynamic Advocate $d_j^{+}$ constructs arguments for higher-intensity activities by citing evidence from the analyst reports, and the Static Advocate $d_j^{-}$ counters with evidence for lower-intensity states. The debate history $h_j$ accumulates arguments for downstream synthesis. In practice, $R{=}2$ rounds suffice for convergence in 85\% of cases.

\textbf{Stage 3: Decision Synthesis.} The Decision Agent acts as the final arbiter, evaluating the accumulated debate history $h_R$ against the primary evidence from analyst reports. It synthesizes these inputs to produce the final classification $\hat{y}$, a quantitative confidence score $\hat{c}$, and a natural language reasoning chain $\mathcal{E}$ for enhanced interpretability.

\vspace{-0.10in}

\section{Evaluation}

The evaluation of SensingAgents is designed to rigorously assess its performance and robustness across multiple dimensions.

\vspace{-0.10in}

\subsection{Dataset and Experimental Setup}
Our evaluation utilizes the widely recognized Shoaib dataset \cite{shoaib2014fusion}, a publicly available benchmark specifically designed for multi-sensor HAR. This dataset captures IMU data from ten participants performing seven distinct daily activities: Walking, Jogging, Sitting, Standing, Biking, Upstairs, and Downstairs. The data was recorded using smartphones strategically placed at five body positions: right jeans pocket, left jeans pocket, belt, right upper arm, and right wrist, capturing 3D accelerometer and gyroscope data at a 50 Hz sampling rate.

To prepare the data, we segmented the continuous IMU streams into 2-second windows with a 50\% overlap. We established several robust baseline models for a comprehensive comparison:
\begin{enumerate}
    \item \textbf{Traditional Machine Learning (SVM)} \cite{hearst1998support}: A supervised machine learning model grounded in the principle of maximum margin, designed to analyze data for classification and regression tasks.
    \item \textbf{Deep Learning (IMU-BERT)} \cite{xu2021limu}: A state-of-the-art deep learning model, which adapts the BERT architecture for IMU data. It learns robust representations to boost HAR performance with limited labels.
    \item \textbf{Single-Agent (with original data)}: A single LLM agent tasked with processing the unoptimized multi-sensor time-series data directly.
    \item \textbf{Single-Agent LLM (based on HARGPT \cite{ji2024hargpt})}: A single LLM agent processing downsampled and statistical sensor data with designed tools.
    \item \textbf{Multi-Agent (Code Generation \cite{claudecode, openclaw})}: An LLM-generated multi-agent framework that attempts to solve the task by generating and executing code without human-expert design.
\end{enumerate}

SensingAgents was implemented using LangGraph with Claude Sonnet 4.6 as the foundation model for all agents. All LLM-based methods were evaluated under zero-shot conditions (no task-specific training data).

\vspace{-0.10in}

\subsection{Overall Performance Comparison}

\begin{figure}[!h]
  \centering
  \includegraphics[width=0.8\linewidth]{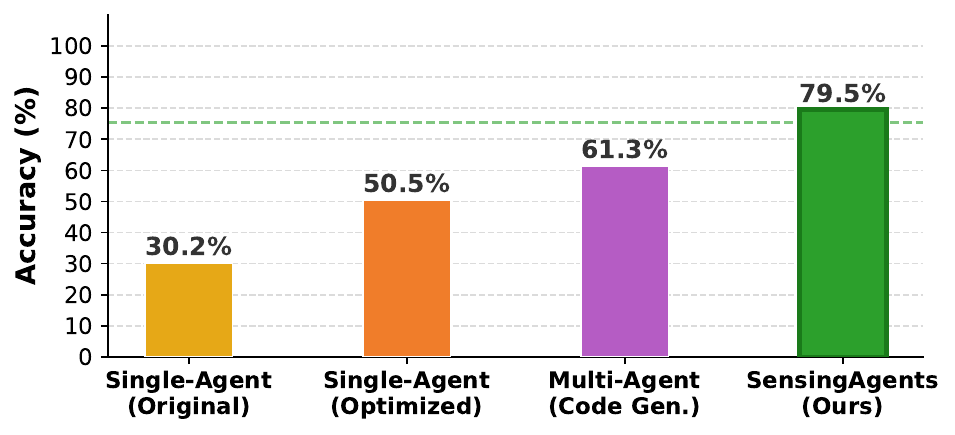}
  \vspace{-10pt}
  \caption{Overall classification performance across agent-based models.}
  \label{fig:overall_performance}
  \vspace{-2.5pt}
\end{figure}

The overall performance is summarized in Figure~\ref{fig:overall_performance} and Table~\ref{tab:unseen_performance}. Within the zero shot regime, a clear performance hierarchy emerges among LLM based methods. The Single-Agent LLM with original data achieves only 30.2\%, confirming that current LLMs cannot reliably interpret raw, high frequency IMU time series. Optimized Single-Agent based on HARGPT, via downsampling and statistics, raises accuracy to 50.5\%, yet forcing a single model to reason over all sensor positions simultaneously still induces cognitive overload and hallucinations. The Multi-Agent (Code Generation) baseline improves further to 61.3\% through multi-agent decomposition, but its LLM generated signal processing pipelines frequently produce unreliable code, including infinite loops and statistically flawed feature extractors, underscoring the necessity of human designed domain tools.

\definecolor{highlightgreen}{rgb}{0.85, 0.95, 0.85}
\definecolor{highlightred}{rgb}{0.95, 0.85, 0.85}

\begin{table}[htbp]
\caption{Performance comparison with supervised learning methods under different conditions.}
\label{tab:unseen_performance}
\centering
\setlength{\tabcolsep}{2pt}
\renewcommand{\arraystretch}{1.2}  
\begin{tabular}{lccc}
\toprule
\textbf{Model / Method} & \textbf{Seen Users} & \textbf{Unseen Users} & \textbf{Decline} \\
\midrule
SVM & 85.2\% & 71.3\% & -13.9\% \\
IMU-BERT & 98.4\% & 70.1\% & -28.3\% \\
\midrule  
\rowcolor{highlightgreen}
\textbf{SensingAgents (Ours)} & ——\% & \textbf{79.5\%} & \textbf{0\%} \\
\bottomrule
\end{tabular}
\end{table}

SensingAgents achieves 79.5\% accuracy, outperforming the best single-agent baseline by around 40 percentage points. Two design principles underpin this gain. First, expert designed signal processing tools, such as FFT analysis, step detection, and vertical asymmetry, offload numerically demanding computations from the LLM, providing compact feature descriptors that ground reasoning in physical quantities. Our ablation confirms that tool calling alone contributes approximately 20 percentage points. Second, position-specific analyst decomposition paired with dialectical debate enables principled resolution of inter-sensor conflicts that confound monolithic architectures.

On the other hand, Table~\ref{tab:unseen_performance} reveals a generalization advantage of SensingAgents. In this evaluation, we use data from a single user for training and treat the remaining nine users as unseen users to assess cross-user generalization. Although supervised models achieve the highest absolute accuracy with labeled training data and seen user condition, SVM and IMU-BERT suffer 13.9\% and 28.3\% accuracy decline on unseen users, respectively, because their learned representations overfit to participant-specific motion patterns present in training data. All zero-shot LLM methods, including SensingAgents, maintain identical performance across seen and unseen users. SensingAgents achieves the highest absolute accuracy (79.5\%), establishing it as the most robust zero-shot approach. It requires no task-specific training, and its modular architecture is designed to benefit directly from improvements in future foundation models without architectural modification.

\vspace{-0.10in}

\subsection{Per-Class Classification Analysis}



\begin{figure}[!h]
  \centering
  \includegraphics[width=0.75\linewidth]{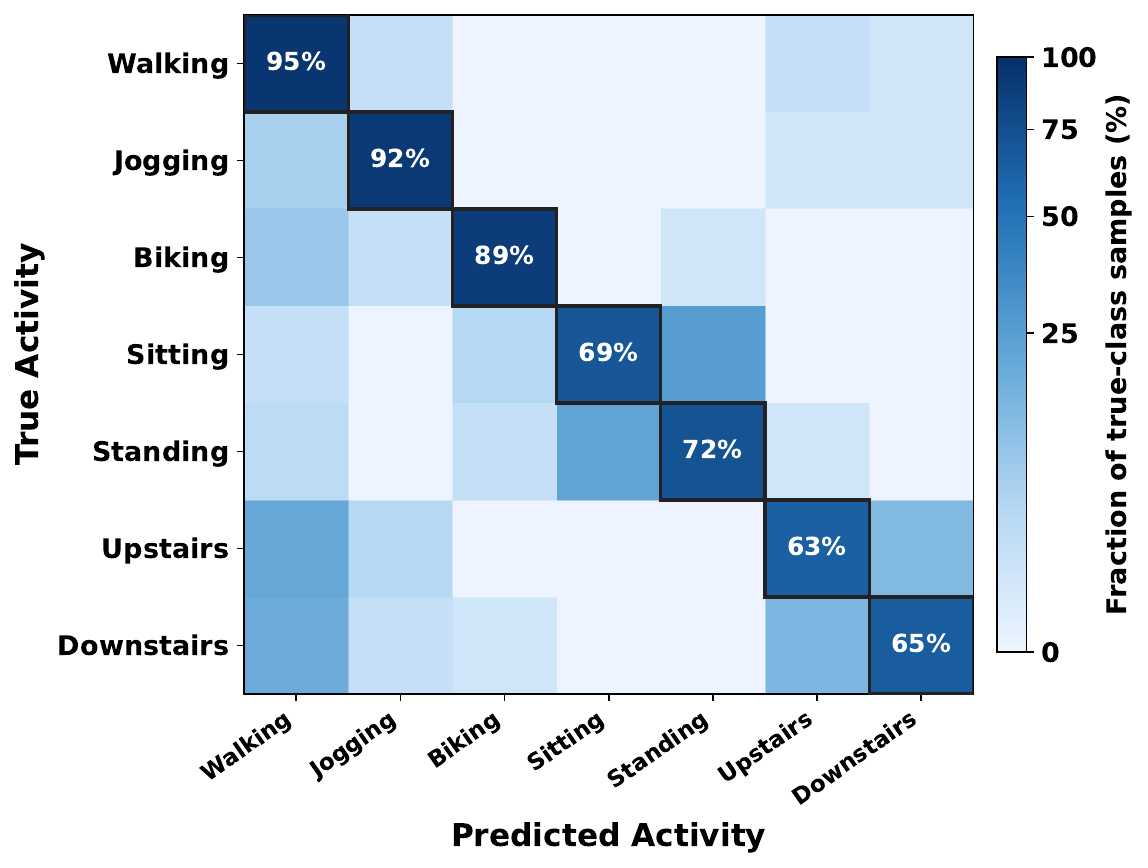}
  \vspace{-10pt}
  \caption{Confusion matrix of SensingAgents. Darker diagonal cells indicate higher per-class accuracy; off-diagonal mass reveals systematic confusion pairs.}
  \label{fig:confusion_matrix}
  \vspace{-2.5pt}
\end{figure}

Figure~\ref{fig:confusion_matrix} shows the full confusion matrix across seven activities, revealing three failure modes that per-class accuracy masks.

\noindent\textbf{Observation 1 --- Static postural confusion (Sitting $\leftrightarrow$ Standing).}
These activities form the largest off-diagonal cluster (~25\% and 22\% mutual error) due to shared near-zero accelerometer and gyroscope variance. Disambiguation relies on subtle gravity vector shifts ($a_z$ vs. $a_y$), which are highly sensitive to device orientation. This lack of distinct kinematic evidence leads to low-confidence draws during agent arbitration.

\noindent\textbf{Observation 2 --- Stair activities misidentified as level-ground walking.}
Upstairs and Downstairs frequently leak toward Walking (20\% and 18\%) because all three share similar rhythmic cadences (1.8–2.2 Hz). The evaluation windows are too short for Tier-3 asymmetry tools like acceleration skewness ($\gamma_y$) to reliably distinguish them, a problem compounded by inconsistent sensor positioning.

\noindent\textbf{Observation 3 --- Stair direction confusion (Upstairs $\leftrightarrow$ Downstairs).}
A secondary error cluster (13–14\%) exists between stair directions. While positive vs. negative skewness should theoretically differentiate upward pushes from downward impacts, these estimates remain unstable across short windows, as asymmetry signatures invert depending on the gait cycle sampling.

Dynamic activities (Walking 95.1\%, Jogging 91.7\%, Biking 88.5\%) are reliable due to strong periodic signatures, negligible cross-class confusion. These findings establish a clear priority ordering for improvement: postural disambiguation requires orientation-invariant features, stair-vs-walking separation requires longer evaluation windows, and stair-direction discrimination requires more precise temporal alignment of the asymmetry analysis window.

\vspace{-2.5pt}
\begin{table}[htbp]
\caption{Performance across different LLM Foundation Models}
\vspace{-10pt}
\label{tab:llm_models_performance}
\centering
\begin{tabular}{lc}
\toprule
\textbf{Foundation Model} & \textbf{Accuracy} \\
\midrule
GPT-5-mini & 74.1\% \\
Deepseek-chat & 73.4\% \\
MiniMax-2.7 & 78.6\% \\
GLM-4.7 & 77.9\% \\
Claude-Sonnet-4.6 & \textbf{79.5\%} \\
\bottomrule
\end{tabular}
\end{table}
\vspace{-2.5pt}

\vspace{-0.10in}

\subsection{Impact of LLM Foundation Models}

To understand how different foundation models affect framework performance, we evaluated SensingAgents using various LLMs as the reasoning engine (Table~\ref{tab:llm_models_performance}). SensingAgents maintains strong performance across all evaluated models, with accuracies from 73.4\% (Deepseek-chat) to 79.5\% (Claude Sonnet 4.6), confirming that architectural design provides consistent gains regardless of the specific model. The five models span a balanced range of providers, scales, and cost tiers, enabling practical trade-off analysis between accuracy and inference cost. A general trend shows models with stronger reasoning capacity and larger scale perform better (MiniMax-2.7, GLM-4.7, Claude Sonnet-4.6). However, the performance spread is narrow (approximately 6 percentage points), indicating SensingAgents provides a substantial and consistent performance floor. We attribute this to the signal processing toolbox offloading numerically demanding computations from the LLM, allowing even moderately capable models to reason effectively over compact semantic descriptors.

Consequently, a natural performance ceiling emerges: once an LLM reliably interprets feature summaries and follows ReAct reasoning loops, further model scaling yields diminishing returns. Our experiments suggest this ceiling is approximately 80\% under the current design, pointing to tool quality and dataset complexity as the primary bottlenecks rather than model capability alone. Claude Sonnet 4.6 achieves the highest average accuracy across multiple independent runs. Nevertheless, for inference latency and cost efficiency, Deepseek-chat presents a compelling alternative, delivering competitive performance at substantially lower cost per token, making it a practical choice for high-throughput or resource-constrained deployments where the modest accuracy trade-off is acceptable.

\begin{figure*}[!t]
  \centering
  \begin{minipage}{0.32\linewidth}
    \centering
    \includegraphics[width=\linewidth]{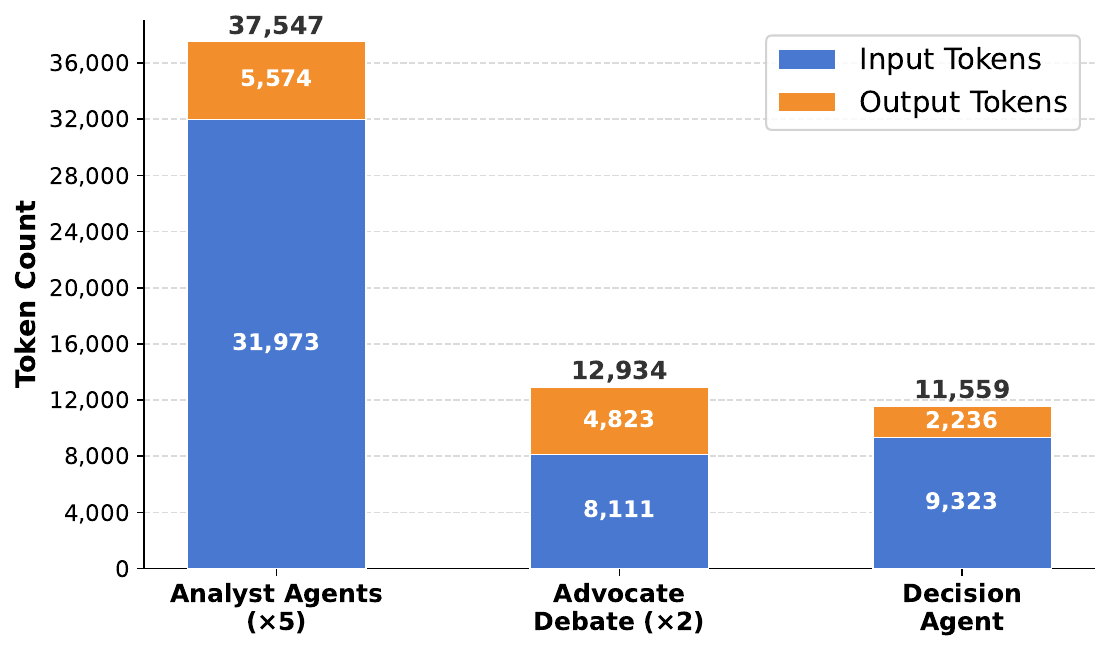}
    \caption{Average token consumption per sample, broken down by pipeline stage and token type (input vs.~output).}
    \label{fig:token_consumption}
  \end{minipage}
  \hfill
  \begin{minipage}{0.32\linewidth}
    \centering
    \includegraphics[width=\linewidth]{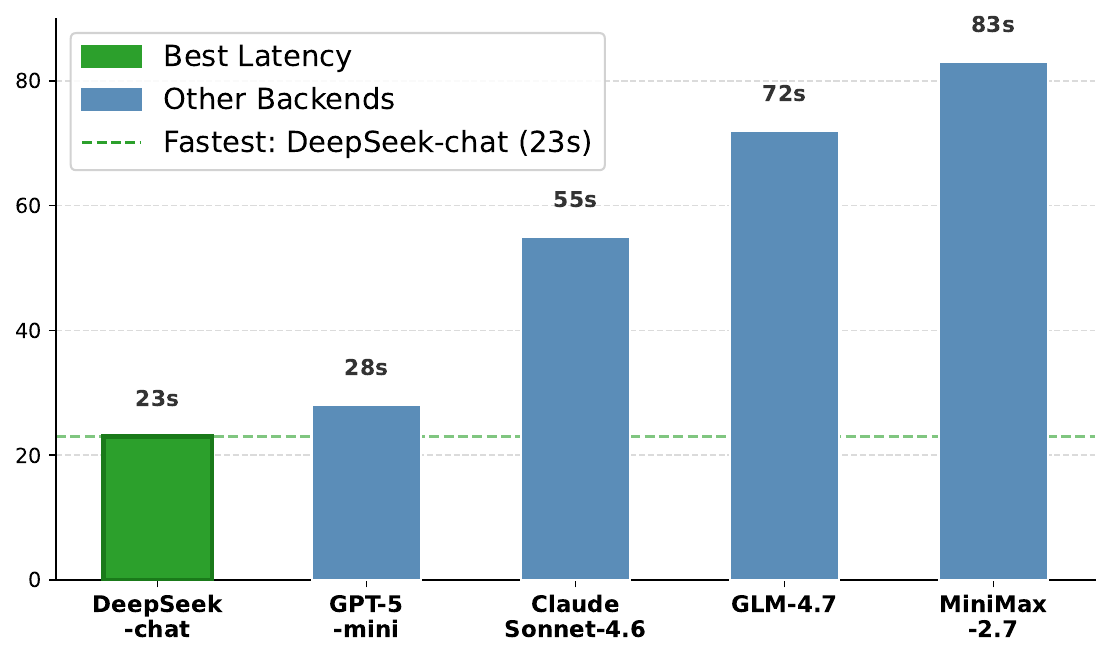}
    \caption{Average end-to-end inference time per sample across five LLM backends.}
    \label{fig:inference_speed}
  \end{minipage}
  \hfill
  \begin{minipage}{0.32\linewidth}
    \centering
    \includegraphics[width=\linewidth]{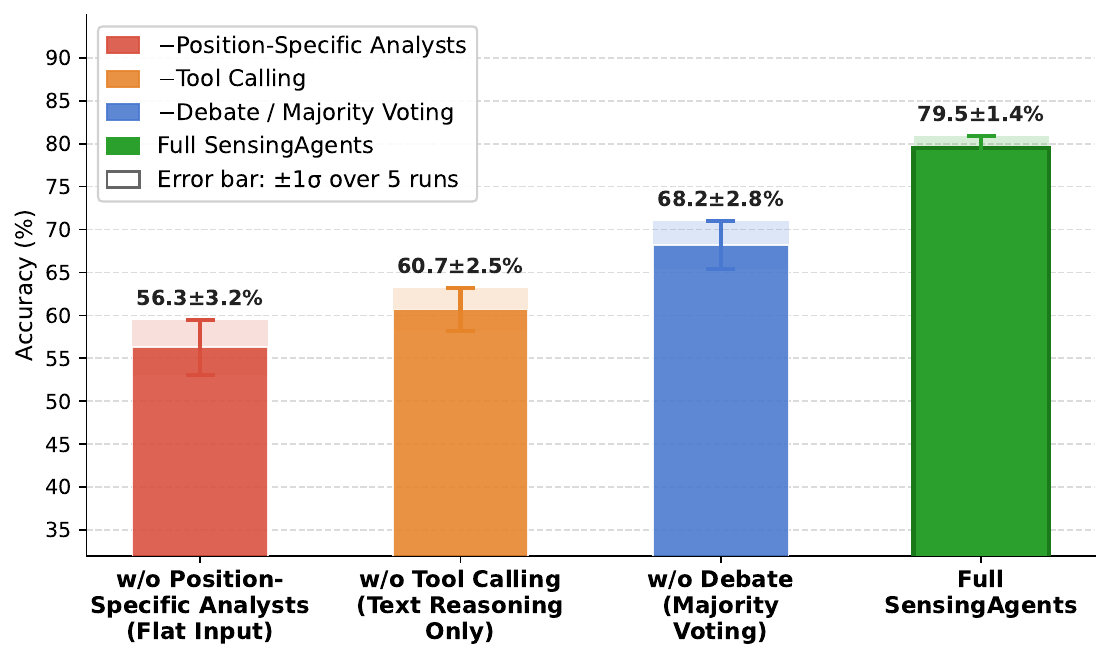}
    \caption{Ablation study of SensingAgents components, includes error bars to quantify stability.}
    \label{fig:ablation_study}
  \end{minipage}
\end{figure*}

\vspace{-0.15in}
\subsection{Token Consumption Analysis}
A practical concern for LLM-based sensing systems is inference cost, which scales with the number of tokens consumed per classification. Figure~\ref{fig:token_consumption} reports token usage per sample. SensingAgents averages 62,040 tokens, with 49.4K input and 12.6K output tokens. Analysts account for 60.5\% of consumption, while the Debate and Decision agents use 20.8\% and 18.6\% respectively.

Two structural insights emerge from this distribution. \textbf{Insight 1: Costs are prompt-bound.} The 4:1 input-to-output ratio indicates that expenditure is driven by fixed overhead like system prompts. This is especially clear in the analyst stage (5.7:1) compared to the debate stage (1.7:1), where agents generate more reasoning. Consequently, prompt compression and schema optimization are the primary levers for cost reduction. \textbf{Insight 2: Debate efficiency.} The Advocate Debate is highly efficient, delivering a 11.3\% accuracy boost for only 20.8\% of the token budget (ablation study). While analysts represent a necessary architectural cost, the debate stage offers the highest return on investment. Refining debate prompts with targeted cues provides a clear path for future optimization.

Compared to a 15,000-token Single-Agent baseline, SensingAgents uses 4x more tokens but increases accuracy from 50.5\% to 79.5\%. This non-linear relationship reflects a qualitative shift in reasoning: from more LLM calls to decomposed position-local reasoning with deterministic signal processing. The modular architecture also reduces cost—skipping analysts for unavailable or unreliable sensors cuts token consumption proportionally without architectural changes.

\vspace{-0.10in}

\subsection{Inference Speed Analysis}


Figure~\ref{fig:inference_speed} reports the average end-to-end inference latency per sample across five LLM backends, measured as the total wall-clock time for all pipeline stages to complete. DeepSeek-chat achieves the lowest latency at 23\,s per sample, making it the most practical backend for latency-sensitive deployments. GPT-5-mini follows at 28\,s, while Claude Sonnet-4.6, GLM-4.7, and MiniMax-2.7 require 55\,s, 72\,s, and 83\,s respectively, primarily due to higher per-token generation latency. Notably, the latency-accuracy trade-off is significant. While Claude Sonnet-4.6 yields the highest accuracy, DeepSeek-chat offers a $3\times$ speed advantage with competitive accuracy, making it the most practical choice for resource-constrained or real-time deployments. In practice, inference latency can be substantially reduced through three complementary strategies: (i)~\textit{Token budgeting}---constraining system prompt length and feature descriptor verbosity; and (ii)~\textit{Adaptive debate truncation}---skipping the second debate round when the first round achieves high-confidence consensus. These optimizations are particularly relevant for edge deployments.

\vspace{-0.10in}

\subsection{Ablation Study}


To validate the contribution of each architectural component within SensingAgents, we conducted an ablation study by systematically removing key features.

The results reveal a clear hierarchy of component importance. Position-specific analysts deliver the largest performance gain, improving accuracy by 23.2 percentage points over flat input, confirming that LLMs benefit fundamentally from decomposed, position-local reasoning. Tool-calling provides the second-largest gain of 18.8 percentage points, validating our core hypothesis that deterministic signal processing tools are essential for grounding LLM reasoning in precise numerical features. The dialectical debate contributes an additional 11.3 percentage points over naive majority voting, demonstrating its value in resolving ambiguous cases where analyst reports conflict. The full SensingAgents model achieves 79.5\% accuracy, substantially outperforming all ablated variants with the lowest standard deviation (1.4), indicating consistent and robust performance.


\vspace{-0.10in}

\section{Discussion}
In this section, we discuss current limitations and outline directions for future work. First, our evaluation is restricted to a single dataset collected under controlled sensor placements. As a result, generalization to free-living scenarios with arbitrary device orientations and missing sensors remains an open challenge. To address this, we plan to extend our evaluation to self-collected datasets and assess cross-domain robustness. Second, the current tool design depends heavily on human domain expertise. While effective, this reliance may limit scalability. A promising direction is automated tool synthesis, where the LLM proposes new feature extraction functions and validates them on a held-out calibration set, potentially uncovering more discriminative features beyond the existing toolbox. Finally, post-training strategies based on agentic reinforcement learning offer another avenue for improvement. By optimizing the debate and decision agents using classification feedback, such methods may further narrow the gap with fully supervised approaches.

\vspace{-0.10in}

\section{Conclusion}
We presented SensingAgents, a multi-agent collaborative framework that decomposes multi-position IMU activity recognition into three specialized layers: position-specific analyst agents with domain-expert tool-calling, adversarial Dynamic-Static advocate agents for conflict resolution, and a decision agent for reliability inference. On the Shoaib benchmark, SensingAgents achieves 79.5\% accuracy in a zero-shot setting, outperforming single-agent LLM baselines by over 50 percentage points and supervised models under unseen conditions by 9 percentage points. Our ablation study confirms that each architectural component contributes measurably to the final performance. These results demonstrate that structured multi-agent reasoning can effectively bridge the modality gap between language models and sensor data, opening new directions for trustworthy, interpretable AI in ubiquitous sensing.

\bibliographystyle{ACM-Reference-Format}
\bibliography{references}

\end{document}